\documentclass[runningheads]{llncs}

\usepackage{eccv}

\usepackage{eccvabbrv}

\usepackage{graphicx}
\usepackage{booktabs}
\usepackage[accsupp]{axessibility} 

\usepackage{enumitem}
\usepackage{svg}
\usepackage{multirow}
\usepackage{float}

\usepackage{hyperref}

\usepackage{orcidlink}

\begin{document}

\title{DriveDiTFit: Fine-tuning Diffusion Transformers for Autonomous Driving}

\author{Jiahang Tu\inst{1} \and
  Wei Ji\inst{2} \and
  Hanbin Zhao\inst{1}\thanks{Corresponding Author. Code is available at: \url{https://github.com/TtuHamg/DriveDiTFit}} \and
  Chao Zhang \inst{1} \and \\
  Roger Zimmermann\inst{2} \and
  Hui Qian\inst{1} }

\authorrunning{Tu et al.}

\institute{Zhejiang University, Zhejiang Province, 310007, China \and
  National University of Singapore, 119077, Singapore}

\maketitle

\begin{abstract}
  In autonomous driving, deep models have shown remarkable performance across various visual perception tasks with the demand of high-quality and huge-diversity training datasets. Such datasets are expected to cover various driving scenarios with adverse weather, lighting conditions and diverse moving objects. However, manually collecting these data presents huge challenges and expensive cost. With the rapid development of large generative models, we propose DriveDiTFit, a novel method for efficiently generating autonomous \textbf{Driv}ing data by \textbf{Fi}ne-\textbf{t}uning pre-trained \textbf{Di}ffusion \textbf{T}ransformers (DiTs). Specifically, DriveDiTFit utilizes a gap-driven modulation technique to carefully select and efficiently fine-tune a few parameters in DiTs according to the discrepancy between the pre-trained source data and the target driving data. Additionally, DriveDiTFit develops an effective weather and lighting condition embedding module to ensure diversity in the generated data, which is initialized by a nearest-semantic-similarity initialization approach. Through progressive tuning scheme to refined the process of detail generation in early diffusion process and enlarging the weights corresponding to small objects in training loss, DriveDiTFit ensures high-quality generation of small moving objects in the generated data. Extensive experiments conducted on driving datasets confirm that our method could efficiently produce diverse real driving data.

\end{abstract}

\section{Introduction}
\label{sec:intro}
\begin{figure}[tb]
  \centering
  \setlength{\abovecaptionskip}{0.cm}
  \includegraphics[height=5.5cm]{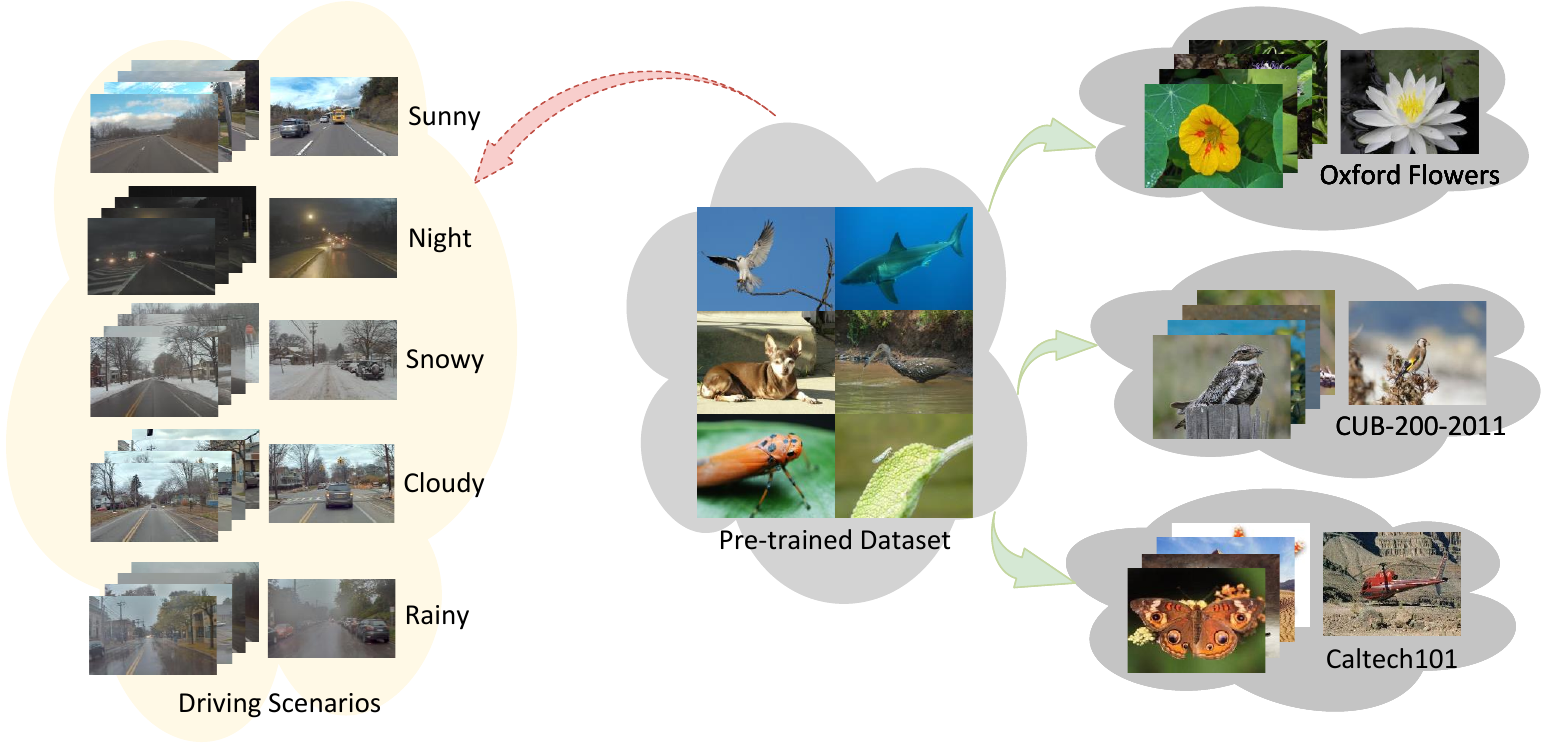}
  \caption{There is an apparent discrepancy between pre-trained datasets and driving scenario datasets. Pre-trained datasets usually feature certain categories of objects prominently displayed within the images, which is similar to the fine-grained classification datasets, such as CUB-200-2011 and Oxford Flowers. However, driving scenario datasets are more complex and contain multiple objects, including roads, vehicles and buildings, with diverse weather and lighting conditions.}
  \label{fig:explanation}
  \vspace{-15pt}
\end{figure}

Recent years have witnessed a rapid development of deep learning models on the autonomous driving application\cite{ad_ap1,ad_ap2}. The performance of data-driven deep models often corresponds to the quality and diversity of the driving data, which necessitates the construction of high-quality and huge-diversity training datasets\cite{bdd100k,ithaca365,zenseact}. Specifically, such datasets are expected to cover various driving scenarios with adverse weather, lighting conditions and diverse moving objects. As demonstrated in \cref{fig:explanation}, in addition to sunny and day scenarios, driving datasets should contain snowy, rainy and night scenarios with vehicles in the field of view. However, manually collecting these data are challenging and expensive, especially for data containing extreme weather scenarios and small moving objects. Due to the outstanding capabilities of Large-scale Generative Models (LGM)\cite{sdxl, dalle, sd3}, this paper focuses on generating data efficiently with LGM to further promote the development of deep models on autonomous driving.

Directly training a LGM from scratch for driving scenarios is time-consuming and resource-expensive, some of recent works try to fine-tune a pre-trained LGM on specific downstream tasks. For example, Xie \etal \cite{difffit}propose to fine-tune biases in diffusion transformers (DiTs) and validate the efficiency on commonly used fine-grained natural classification datasets. Hence, we focus on fine-tuning a pre-trained LGM efficiently to adapt to the complex autonomous driving scenarios. The LGM is always pre-trained on a natural classification dataset (\eg, ImageNet\cite{imagenet}) and encodes the class-specific knowledge with a class embedding module. As depicted in \cref{fig:explanation}, such dataset mainly includes a prominent natural-category object located in the central position of the image, but driving data usually contains vehicles with diverse weather and lighting conditions. Motivated by these observations, we consider the discrepancy and aim to design a parameter-efficient LGM fine-tuning method tailored for autonomous driving scenarios.

\begin{figure}[tb]
  \centering
  \includegraphics[height=5cm]{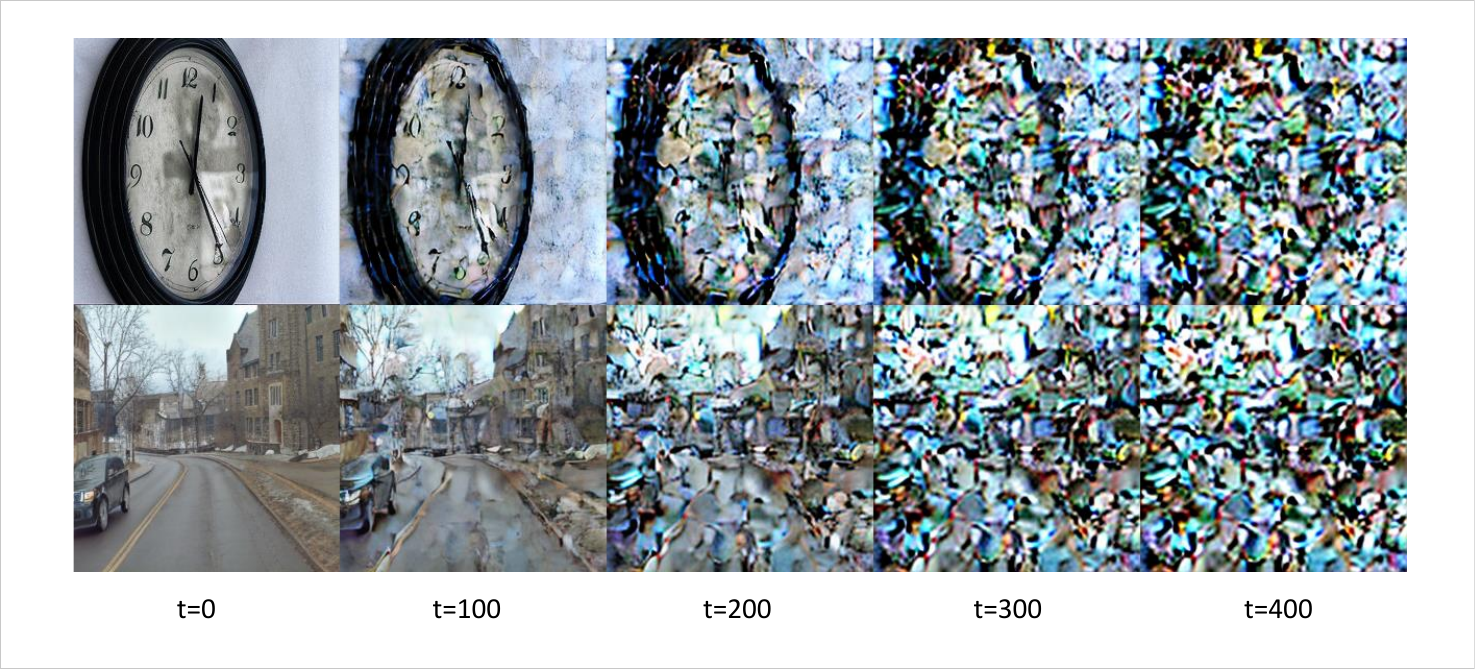}
  \caption{
    The object information can be generated sufficiently in the denoising process\cite{simple}, when it loses slowly in the diffusion process.
    The conventional noise schedule\cite{dit} makes big objects in classification dataset lose slowly (top row, clock, t from 0 to 400), whereas it causes smaller objects in driving data to fade more rapidly (bottom row, vehicles, t from 0 to 200). An appropriate noise schedule is necessary for driving data generation.
  }
  \label{fig:noise_concat}
  \vspace{-15pt}
\end{figure}

In this paper, we propose DriveDiTFit to efficiently generate high-quality and huge-diversity autonomous \textbf{Driv}ing data by \textbf{Fi}ne-\textbf{t}uning pretrained \textbf{Di}ffusion \textbf{T}ransformers. Specifically, DriveDiTFit utilizes a gap-driven modulation scheme to fine-tune only a few parameters of the pretrained DiTs. To generate adverse weather and lighting conditions, DriveDiTFit effectively embeds the weather and lighting condition-specific knowledge by an expandable condition embedding module, which is initialized by a nearest-semantic-similarity initialization approach. Besides, we observed that the noise schedule in the diffusion process\cite{simple,iddpm} can greatly affect the quality of objects in the generated data as depicted in \cref{fig:noise_concat}. Our DriveDiTFit designs a spoon-cosine (Scos) noise schedule and progressively adjusts the noise intensity to mitigate the impact on original parameters. To further ensure the generated quality of small objects, we explicitly introduce the extra position knowledge of small objects and develop an object-sensitive loss function by enlarging the weights corresponding to the regions of the small objects.
Overall, the contribution of our work are three-fold:

\begin{itemize}[topsep=0pt]
  \item[\textbullet] We propose a gap-driven modulation technique for parameter-efficient fine-tuning DiT models to address the large gap between source datasets and target datasets.
  \item[\textbullet] We present a semantically relevant embedding initialization method, leveraging the prior knowledge to embed the weather and lighting conditions. This approach enhances tuning convergence and improves the generation quality.
  \item[\textbullet] To enhance the detail fidelity of generated small objects in driving scenarios, we implement a progressive tuning scheme with an innovative Scos noise schedule and introduce a loss function sensitive to small objects.
\end{itemize}

\section{Related Work}

\subsection{Autonomous Driving Datasets}
The KITTI dataset\cite{kitti}, widely utilized in research on driving algorithms, serves as a benchmark for tasks like 2D object detection and optical flow. It features a rich collection of RGB, LiDAR, and GPS/IMU data, albeit predominantly under daytime and clear weather conditions. To broaden the scope of environmental conditions, the Waymo\cite{waymo} and Zenseact\cite{zenseact} datasets cover night scenarios and introduce dawn and dusk scenarios, respectively, and the nuScenes\cite{nuscenes} dataset expands coverage to rainy weather but lacks snow scenes. Notably, the BDD100K\cite{bdd100k} dataset collects five types of weather conditions including clear, cloudy, rainy, snowy, and foggy settings. As shown in \cref{tab:headings}, given the unbalanced diversity in conditions, the predominance of clear weather and daylight scenarios in these datasets often leads to decreased performance of visual models under nighttime or adverse weather conditions. Ithaca365\cite{ithaca365} collects repeated trajectories under diverse scene, weather, time and traffic conditions over 1.5 year period but its data scale is smaller compared to the previously mentioned datasets. The manual acquisition of driving datasets from real-world scenarios is both time-consuming and labor-intensive.

\begin{table}[tb]
  \setlength{\tabcolsep}{5pt}
  \caption{Statistical comparisons of weather and lighting conditions in driving datasets.
  }
  \label{tab:headings}
  \centering
  \begin{tabular}{@{}c|cccc|ccc@{}}
    \toprule
    \multirow{2}{*}{Driving datasets} & \multicolumn{4}{|c|}{Weather conditions(\%)} & \multicolumn{3}{c}{Time of day(\%)}                                                                 \\
                                      & \multicolumn{1}{|c}{clear/cloudy}            & rainy                               & snowy & \multicolumn{1}{c|}{foggy} & day  & night & dawn/dusk \\
    \midrule
    KITTI\cite{kitti}                 & 100                                          & -                                   & -     & -                          & 100  & -                 \\
    nuScenes\cite{nuscenes}           & 78.1                                         & 21.9                                & -     & -                          & 88.4 & 11.6  & -         \\
    Waymo\cite{waymo}                 & 99.4                                         & 0.6                                 & -     & -                          & 81.0 & 9.9   & 9.1       \\
    Zenseact\cite{zenseact}           & 80.2                                         & 15.7                                & 2.0   & 2.1                        & 77.3 & 19.0  & 3.6       \\
    BDD100K\cite{bdd100k}             & 66.0                                         & 6.9                                 & 7.8   & 0.1                        & 52.6 & 40.4  & 7.2       \\
    \bottomrule
  \end{tabular}
\end{table}

\subsection{Diffusion Models}

Diffusion probabilistic models (DPMs)\cite{ddpm} have emerged as a powerful class of generative models, surpassing outperform GANs\cite{gan} and  variational autoencoders (VAEs)\cite{vae,pvi} in a variety of tasks, including text-to-image generation\cite{ldm,adma,sinkhorn}, image editing\cite{sdedit, apisr} and video synthesis\cite{synthetic}. The essence of DPMs lies in incrementally mapping Gaussian noise to intricate distributions related to datasets. Given a certain noise schedule, the diffusion phase converts a data distribution to a standard Gaussian distribution by adding noise. The denoising phase primarily adopts a UNet architecture to iteratively reverse the noise addition, thereby reconstructing the original data distribution. Inspired by breakthroughs in natural language processing\cite{bert} and Vision Transformer (VIT)\cite{vit, pectp}, architectures based on transformers\cite{transformers} have been proposed and replaced the UNet backbone, achieving state-of-the-art results\cite{uvit,dit} on benchmark datasets such as CIFAR10 and ImageNet.

\subsection{Fine-tuning for Diffusion Models}
The emergence of numerous high-quality diffusion models has drawn the attention of researchers, prompting them to explore fine-tuning approaches tailored to their specific requirements. Techniques like Textual Inversion\cite{inversion} and Dreambooth\cite{dreambooth} introduce novel token identifiers for personalized text embedding adjustments, though their application remains largely within text-to-image paradigms and customized datasets are similar to source datasets. DiffFit\cite{difffit} proposed a parameter-efficient method by fine-tuning biases and scale factors in DiTs. Moon \etal\cite{moon} explore the integration of  time-fusion adapters within attention mechanisms on limited datasets. VPT\cite{vpt} freezes transformer blocks and inserts a few learnable prompt embeddings in different layers for downstream tasks tuning. Nevertheless, these works predominantly focus on classification datasets, with limited exploration in the efficiency on scenario datasets. Moreover, the interaction between the original noise schedule and target dataset characteristics remains an underexplored area.

\section{Method}

\subsection{Preliminaries}
Before introducing our novel fine-tuning method for DPMs, we briefly revisit the foundational principles of DPMs. Given a data distribution $x_0\sim q_{data}(x)$, the diffusion phase iteratively adds noise $\epsilon_t$ to the sample $x_t$ until $x_T$, following a certain noise schedule and time $t$. This process can be described as follow:
\begin{align}
  q(x_1,\dots,x_T|x_0) & =\prod_{t=1}^{T}q(x_t|x_{t-1}),                                                         \\
  q(x_t|x_{t-1})       & =\mathcal{N}(x_t;\sqrt{1-\beta_t}x_{t-1},\beta_t\textbf{I}),                            \\
  q(x_t|x_0)           & =\mathcal{N}(x_t;\sqrt{\bar\alpha_t}x_0,(1-\bar\alpha_t)\textbf{I}) \label{eq:x_t|x_0},
\end{align}
where $\alpha_t=1-\beta_t$ and $\bar\alpha_t=\prod_{s=1}^t\alpha_s$. Here, $\beta_t$ represents the noise intensity at each step, and a large number of steps ($T$) enables $x_T$ to approximate a Gaussian distribution closely.

The essence of DPMs lies in their ability to reverse this noise addition process, effectively generating samples from the original data distribution by learning a sequence of reverse mappings. They aim to learn the inverse process $p_\theta(x_{t-1}|x_t)$ and approximate the posterior $q(x_{t-1}|x_t,x_0)$, which are defined as follow:
\begin{align}
  p_\theta(x_{t-1}|x_t) & =\mathcal{N}(x_{t-1};\mu_\theta(x_t,t),\Sigma_\theta(x_t,t)) \label{eq:model},             \\
  q(x_{t-1}|x_t,x_0)    & =\mathcal{N}(x_{t-1};\tilde \mu_t(x_t,x_0),\tilde \beta_t\textbf{I}) \label{eq:posterior},
\end{align}
where $\tilde \mu_t(x_t,x_0)=\frac{\sqrt{\bar{\alpha}_{t-1}}\beta_t}{1-\bar{\alpha}_t}x_0 + \frac{\sqrt{\alpha_t}(1-\bar{\alpha}_{t-1})}{1-\bar{\alpha}_t} x_t$ and $\tilde{\beta}_t=\frac{1-\bar{\alpha}_{t-1}}{1-\bar{\alpha}_t} \beta_t$. Ho \etal\cite{ddpm} hypothesis $\Sigma_\theta$ is not learnable and reparameterize the $ \mu_\theta(x_t,t)$ through \cref{eq:x_t|x_0} and \cref{eq:posterior}:
\begin{align}
  \mu_{\theta}(x_t, t) & = \frac{1}{\sqrt{\alpha_t}} \left( x_t - \frac{\beta_t}{\sqrt{1-\bar{\alpha}_t}} \epsilon_{\theta}(x_t, t) \right).
\end{align}
The simple loss function can be written as follow:
\begin{align}
  \mathcal{L}_{\text{simple}} & = E_{t,x_0,\epsilon_t}\left[ || \epsilon_t - \epsilon_{\theta}(x_t, t) ||^2 \right].
\end{align} During inference, DPMs generate new samples by first initializing $x_T$ from Gaussian distribution. The model then sequentially calculates $x_{t-1}$ through \cref{eq:model} until reaching $x_0$, thus reconstructing or generating new data points that mimic the original data distribution.

\begin{figure}[tb]
  \centering
  \setlength{\abovecaptionskip}{0.cm}
  \includegraphics[height=8.5cm]{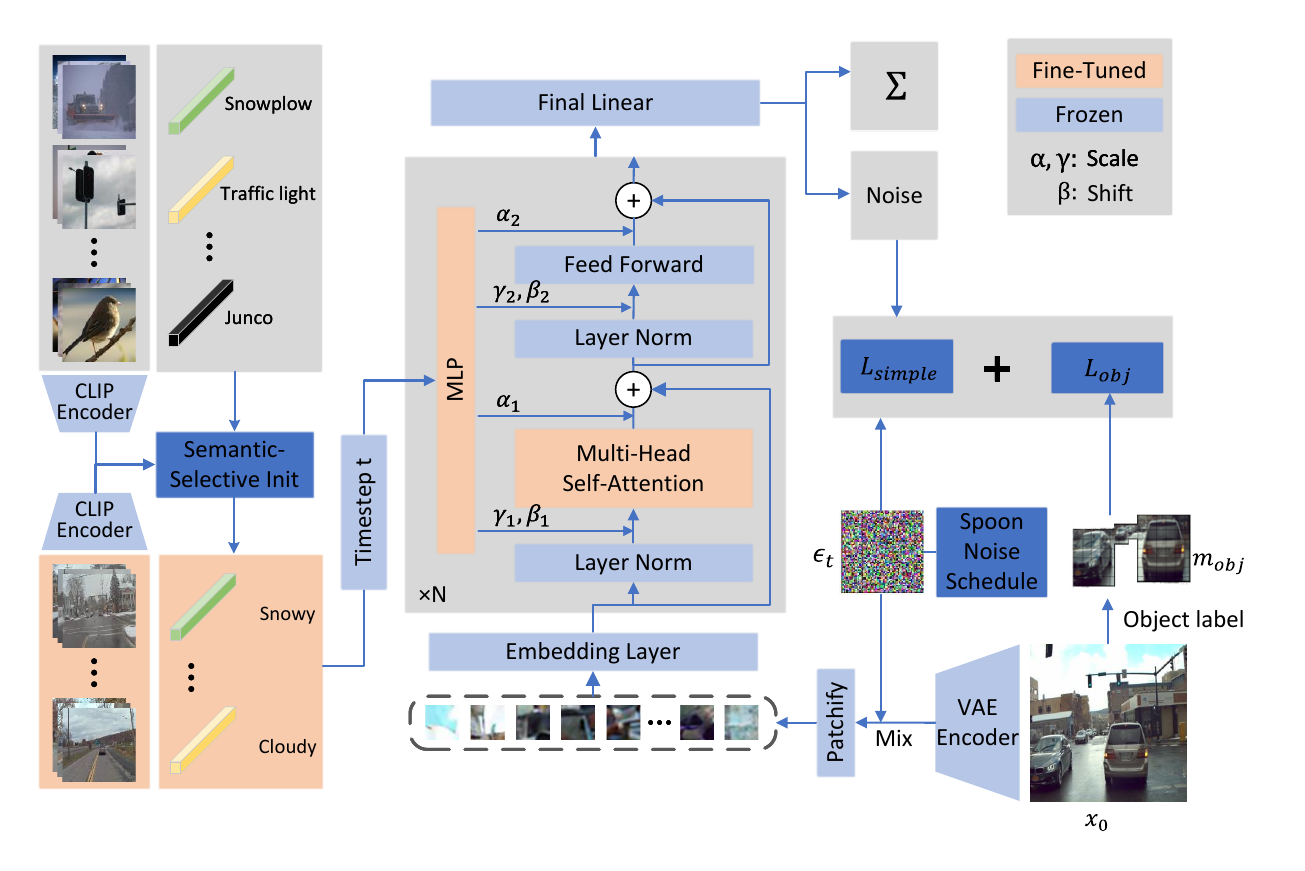}
  \caption{Our framework for diverse driving scenario generation consists of three key components: i) Gap-driven modulation techniques on the condition MLP and attention blocks; (Sec. 3.2); ii) Accelerating convergence and enhancing quality by initiating with high semantic similarity embeddings via a CLIP encoder (Sec. 3.3); iii) Adopting progressive tuning scheme with novel Scos noise schedule  (Sec. 3.4.1) and applying vehicle bounding box masks on training loss (Sec. 3.4.2) for precise object representation.}
  \label{fig:main_fig}
  \vspace{-15pt}
\end{figure}

\subsection{Architecture Fine-tuning}
As previously discussed, driving scenarios present a distinct challenge from traditional classification generation. Specifically, DiTs in driving scenarios must generate images across varied weather and lighting conditions while creating fine details of small objects. This requirement underscores a significant gap: mastering single-category object is insufficient. Instead, models must grasp the global weather and lighting  style and positional relationships with multiple objects. This complexity suggests that methods solely fine-tuning biases and scale factors, as proposed in DiffFit\cite{difffit}, may fall short for driving datasets. Inspired by AdaLN\cite{adaln} in style transfer, which leverages style statistics information to modulate input images, we propose an advanced modulation approach. Specifically, we select different modules in DiTs according to the significant gap. As the changes in conditions and different layout demands of driving datasets, we modulate the weights within the condition MLP and the multi-head self-attention block, employing Low-Rank Adaptation(LoRA) to enhance training efficiency. We obtain the modulated weight $\tilde W\in \mathbb{R}^{d_{out}\times d_{in}}$ like:
\begin{align}
  \tilde W = W \odot \Gamma + B,
\end{align}
where $\Gamma\in \mathbb{R}^{d_{out}\times d_{in}}$ and $B\in \mathbb{R}^{d_{out}\times d_{in}}$ present the fine-tuning parameters. These can be effectively represented by two low-rank matrices:
\begin{align}
  \Gamma = \Gamma^{out} \otimes \Gamma^{in}, B = B^{out} \otimes B^{in},
\end{align}
with $\Gamma^{out}, B^{out}\in \mathbb{R}^{d_{out}\times r}$ and $\Gamma^{in}, B^{in}\in \mathbb{R}^{r\times d_{in}}$. Detailed experiments demonstrate the validity of our fine-tuning scheme on the structure. Furthermore, the generation of local objects in driving scenarios should have a narrower focus compared to category objects in classification datasets. We adopt an approach similar to Peebles and Xie\cite{dit} and introduce 2D rotary positional embeddings (RoPE)\cite{rope}, a technique prevalent in natural language processing, to emphasize local objects generation. For an input $z\in \mathbb{R}^{c\times h\times w}$ in the latent space, we can obtain $\frac{H}{p}\times \frac{W}{p}$ $u\in \mathbb{R}^{c\times p\times p}$ tokens. Applying RoPE along each spatial axis and concatenating, we obtain the position embedding $P_{i,j}\in \mathbb{R}^{d\times d}$:
\begin{align}
  P_{i,j}=
  \begin{bmatrix}
    RoPE(i) & 0       \\
    0       & RoPE(j)
  \end{bmatrix},
\end{align}
where RoPE($\cdot$)$\in \mathbb{R}^{\frac{d}{2}\times \frac{d}{2}}$ and $d$ represents hidden channel number. The local attention mechanism can be written as:
\begin{align}
  (P_{i,j}q_{i,j})^T(P_{i^{\prime},j^{\prime}}k_{i^{\prime},j^{\prime}})=q_{i,j}^TP_{i-i^{\prime},j-j^{\prime}}k_{i^{\prime},j^{\prime}}.
\end{align}
As $P_{i-i^{\prime},j-j^{\prime}}$ is an orthogonal constant matrix, it preserves the magnitude of vectors without adding new learnable parameters to the model, thus ensuring efficiency and model simplicity.

\subsection{Semantic-Selective Embedding Initialization}

The DiT model incorporates learnable embeddings that encapsulate categorical information from classification datasets. Hence, we develop a weather and lighting condition embedding to control diverse scenario generation. Directly initializing the embeddings from scratch is deemed impractical due to the significant impact on parameters. Traditional methods of initializing embeddings through random selection of pre-trained embeddings introduce a certain degree of variability. To address this challenge, we propose a semantic-selective embedding initialization (SSEI) approach. We employ a pre-trained CLIP encoder \cite{clip} to extract semantic features $\{z^{s}_i\}_{i=1}^{N}$ and $\{z^{ad}_j\}_{j=1}^{M}$ from both source datasets  and driving datasets. We then measure the semantic similarity between $z^{s}_i$ and $z^{ad}_j$ with cosine similarity to find an appropriate class embedding $c_{i^*}^{s}$ for certain scenario condition $c_j^{ad}$. The nearest-semantic-similarity embedding initialization can be written as follows:
\begin{align}
  i^* = \mathop{\mathrm{argmax}}\limits_{i} \frac{z_i^s \cdot z_j^{ad}}{\|z_i^s \|_2 \cdot \| z_j^{ad} \|_2}.
\end{align}
Our hypothesis is that image features closely aligned in semantic space are likely to have similar information in DiT embeddings and also exhibit similar distribution characteristics. Through ablation studies, we have demonstrated that our SSEI approach significantly accelerates model convergence and yields superior performance outcomes. This method effectively leverages semantic correlations to enhance the initialization process, setting a robust foundation for better learning and adaptation in driving scenario modeling.

\begin{figure}[tb]
  \centering
  \setlength{\abovecaptionskip}{0.2cm}
  \includegraphics[width=1.0\textwidth]{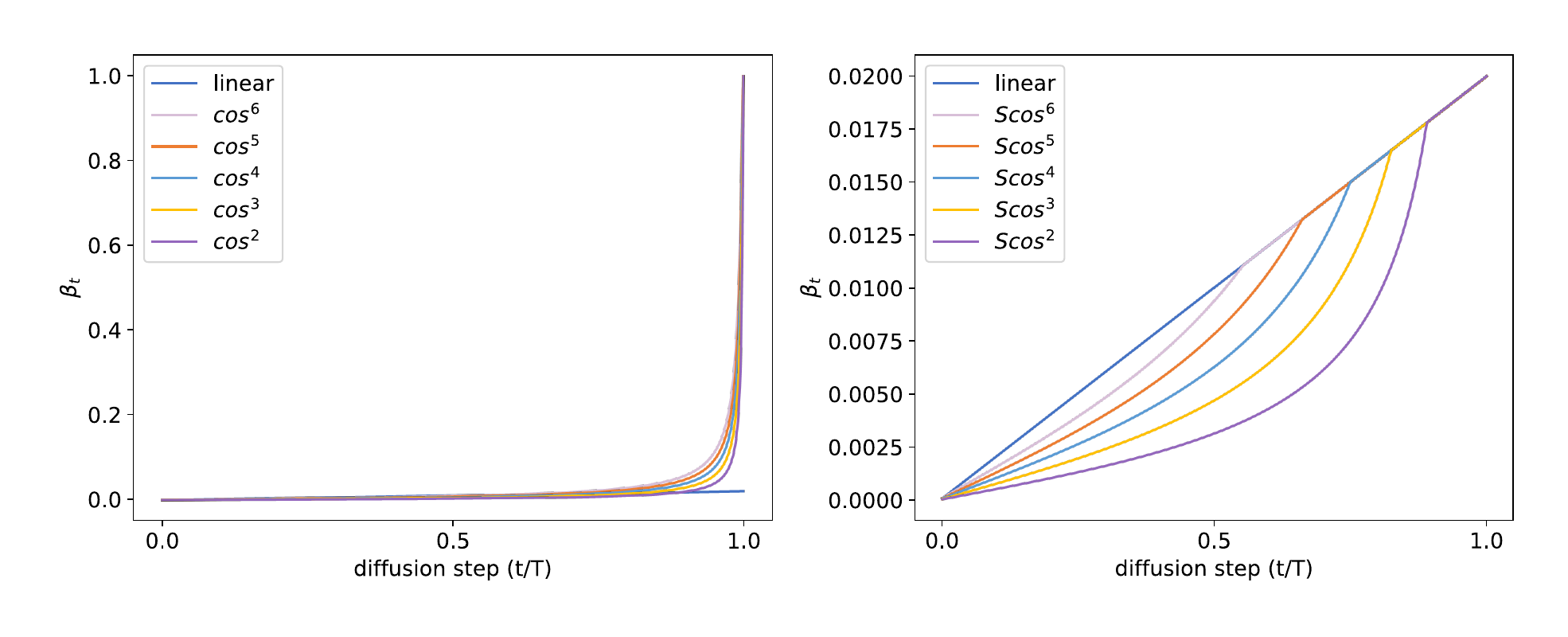}
  \caption{In the diffusion process, $\beta_t$ varies between the cosine noise schedule proposed by Nichol and Dhariwal and Scos noise schedule.}
  \label{fig:bug}
\end{figure}

\subsection{Progressive Tuning Scheme}

\textbf{Spoon-Cosine Noise Schedule.} The conventional linear noise schedule approach, typically employed for classification datasets, proves inadequate for driving datasets. This insight draws inspiration from the work of Hoogeboom \etal\cite{simple} on high resolution images synthesis, which revealed that traditional noise schedules only afford a small time window to decide the global structure of the image at the late diffusion process , retaining good visual quality of high-resolution generation. In contrast, as we demonstrate in \cref{fig:noise_concat}, small objects within scenarios tend to be submerged in the early diffusion process, which lead to a shorter time window for the reverse process of small objects generation. The underlying cause for this phenomenon is that at a resolution of 256$\times$256 pixels, small objects in driving datasets are described by fewer pixels compared to those in classification datasets. Consequently, at equivalent noise intensity, small objects become harder to identify.

\begin{figure}[tb]
  \centering
  \setlength{\abovecaptionskip}{0.0cm}
  \includegraphics[width=1.0\textwidth]{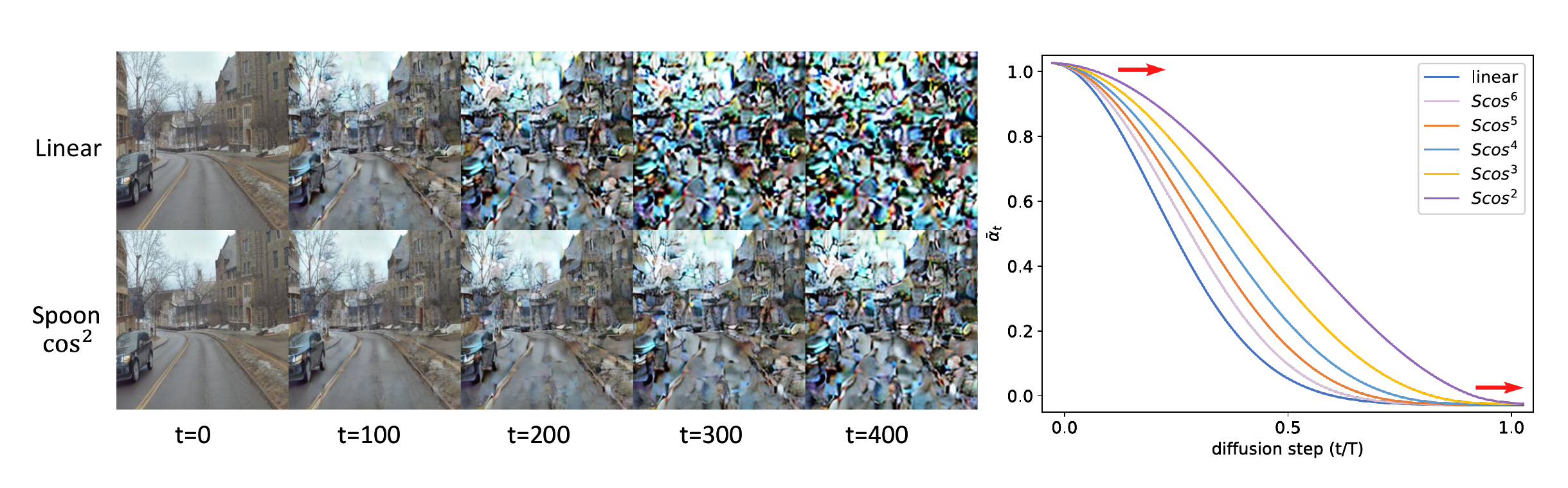}
  \caption{\textbf{Left}: The linear noise schedule(top row) and $\mathrm{Scos^2}$ noise schedule(bottom row) are adopted on an driving scenario sample. \textbf{Right}: $\bar \alpha_t$ in diffusion process for the linear noise schedule and Scos noise schedule with different powers.}
  \vspace{-15pt}
  \label{fig:progressive}
\end{figure}

A naive idea is to alter the original schedule during the fine-tuning, such as employing cosine schedule, which is notably beneficial for driving datasets as it reduces the noise intensity in the early stages of diffusion. However, the practical application of this seemingly straightforward adjustment during fine-tuning presents challenges as it causes great damage to the learned mapping from Gaussian distribution to the dataset distribution. Inspired by Nichol and Dhariwal's\cite{iddpm} squared cosine noise table, we discover that adjusting the power term $s$ effectively captures the gradual transition from a linear to a cosine schedule:
\begin{align}
  f_s(t)        & =\cos^s(\frac{t/T+b}{1+b}\cdot \frac{\pi}{2}), \\
  \bar \alpha_t & =\frac{f_s(t)}{f_s(0)}.
\end{align}
Nevertheless, as demonstrated in \cref{fig:bug}, we observe a significant change in the $\beta_t$ value, which represents noise intensity, at the endpoint of the diffusion process. This change obviously differs from the diffusion patterns learned by pre-trained models. This discovery suggests that the traditional fine-tuning methods on model architecture may not adequately adapt to such changes. Consequently, we comprehensively consider the advantages and disadvantages of the squared cosine noise schedule and propose a novel noise schedule to optimize the adjustment of noise intensity throughout the diffusion process. We have termed this strategy the ``spoon-cosine'' (Scos) noise schedule due to its unique $\beta_t$ value change curve, which resembles the shape of a spoon. As depicted in \cref{fig:progressive}, this noise schedule employs a cosine-like schedule in the early stages of the diffusion process to slow down the loss of details; in the later stages of diffusion, it shifts to the linear schedule to better align with the original noise schedule of pre-trained models. During the fine-tuning process, we progressively tune (PT) the Scos noise schedule from $\mathrm{Scos^6}$ to $\mathrm{Scos^2}$ at time intervals ($\tau$), facilitating the model's ability to generate more intricate and detailed representations.

\textbf{Object-Sensitive Loss.} For high quality objects generation,  particularly of vehicles, our work proposes an object-sensitive loss designed to enhance the precision and quality of generated objects. Specifically, We introduce additional supervision signals into the original loss function, leveraging the bounding box information available within driving datasets. More precisely, we select bounding boxes associated with vehicles and incorporate these prior masks $m_{obj}$ into the loss function by enlarging the weights corresponding to certain regions, which encourages the model to allocate more focus in the generation of vehicles. This object-sensitive loss (OSL) is defined as:
\begin{align}
  \mathcal{L}_{obj}=E_{t,x_0,\epsilon_t}\left[ || m_{obj}(\epsilon_t - \epsilon_{\theta}(x_t, t)) ||^2 \right].
\end{align}

\section{Experiments}
\label{sec:blind}
In this section, we present experiments to validate the effectiveness and motivation of our proposed method. Quantitative and qualitative analysis present DriveDiTFit can generate high-quality and huge-diversity driving data.

\subsection{Implementation}
\textbf{Datasets}
Ithaca365\cite{ithaca365} is an autonomous driving dataset recorded along a 15km route under diverse weather (clear, cloudy, rainy, snowy) and lighting (day, night) conditions. It provides scenario image data with balanced labels across varying scenarios, including urban, highway and rural environments.
BDD100K\cite{bdd100k} is a another substantial driving dataset renowned for its size and diversity. It comprises 70000 images within the training dataset and covers a small amount of foggy scenario data, alongside the aforementioned weather and lighting conditions.

\textbf{Metrics}
We assess the effectiveness of DriveDiTFit in terms of sample quality and the coverage of the data manifold. Fréchet Inception Distance (FID)\cite{fid} leverages pooling features from Inception-V3 to calculate the KL divergence between real and generated samples. Similarly, sFID\cite{sfid} employs spatial features to evaluate the spatial distribution similarly between samples. We report FID and sFID to evaluate the sample quality. Besides, we utilize Improved Precision and Recall metrics\cite{pr} for further evaluation of generative models. Precision quantifies the fraction of generated images that exhibit realism, while recall measures the extent to which the generative models cover the training data manifold. Additionally, we select the object detection downstream task to validate the performance improvement of perception models using the generated dataset. We employ Mean Average Precision (mAP) as the evaluation metric, which represents the average area under the precision-recall curve for all classes in the images.

\begin{table}[tb]
  \setlength{\tabcolsep}{10pt}
  \caption{The performance comparison of DriveDiTFit and other fine-tuning methods on Ithaca365 dataset. Our proposed approach achieves the best generative performance while only requiring fine-tuning 0.35\% of the parameters.
  }
  \label{tab:metric}
  \centering
  \begin{tabular}{@{}lccccc@{}}
    \toprule
    Method                  & FID$\downarrow$   & sFID$\downarrow$  & Prec.$\uparrow$   & Recall $\uparrow$ & Params(M)                 \\ \midrule
    Full Fine-tune          & 37.98             & 31.95             & 0.571             & 0.648             & 675.1(100\%)              \\
    Time-Adapter\cite{moon} & 30.82             & 27.41             & 0.558             & 0.711             & 260.3(38.5\%)             \\
    LoRA-R4\cite{lora}      & 37.38             & 34.05             & 0.425             & 0.649             & 3.000(0.44\%)             \\
    VPT-Deep\cite{vpt}      & 35.32             & 31.24             & 0.358             & 0.677             & 1.380(0.20\%)             \\
    BitFiT\cite{bitfit}     & 29.23             & 31.40             & 0.379             & 0.751             & 0.490(\underline{0.07\%}) \\
    DiffFit\cite{difffit}   & 27.98             & 31.11             & 0.386             & 0.773             & 0.590(0.09\%)             \\ \midrule
    DriveDiTFit             & \underline{18.64} & \underline{26.13} & \underline{0.626} & \underline{0.795} & 2.372(0.35\%)             \\ \bottomrule
  \end{tabular}
\end{table}

\textbf{Details}
We choose DiT-XL/2 as our base models, which were pre-trained on ImageNet at 256x256 resolution over 7 million steps.  To maintain consistency with the pre-trained models, the driving dataset is resized to 256x256 resolution and then divided into five conditions base on weather and lighting properties: sunny, cloudy, rainy, snowy and night. We select DiffFit as our primary comparison method since it has been tested with DiT-XL/2 on a broad range of downstream classification datasets. Additionally, we also re-implement several fine-tuning methods, including Time-Adapter\cite{moon}, Visual Prompt Tuning (VPT)\cite{vpt}, BitFit\cite{bitfit} and LoRA\cite{lora}. For VPT, we seek for a more stable configuration with depth=14, token=3, which is suitable for fine-tuning the driving datasets. In addition to that, other fine-tuning methods adopt zero initialization or identity initialization to alleviate the impact of gradients on parameters. Following DiffFit's tuning setting, we utilize AdamW optimizer with a constant learning rate of 1e-4 for all fine-tuning methods. We use 4 A800 GPUs with a global batch size of 256 for 500 iterations, with a fixed random seed of 0. We select ViT-L-14 as the semantic encoder and set $\tau$ to be equal to the number of training steps. We mainly report the performance of our method on the Ithaca365 dataset. More experimental results can be found in supplementary materials.

\subsection{Quantitative Evaluation}
\textbf{Comparison with the state-of-the-arts.} In this section, we show the quantitative comparison of our proposed DriveDiTFit with other fine-tuning methods. As shown in \cref{tab:metric}, under the condition of an acceptable amount of fine-tuning parameters and comparable other metrics, our proposed DriveDiTFit outperforms DiffFit and BitFit in terms of the FID and Precision metric. This corroborates our initial motivation, suggesting that for fine-tuning tasks with the significant gap in datasets, merely adjusting biases and scaling factors are insufficient. Due to the small size of Ithaca365, tuning more parameters can lead to overfitting, resulting in decreased performance on Full Fine-tuning and Time-Adapter method. The generation samples of different methods are included in the supplementary materials.

\begin{figure}[tb]
  \centering
  \setlength{\abovecaptionskip}{0.2cm}
  \includegraphics[width=1.0\textwidth]{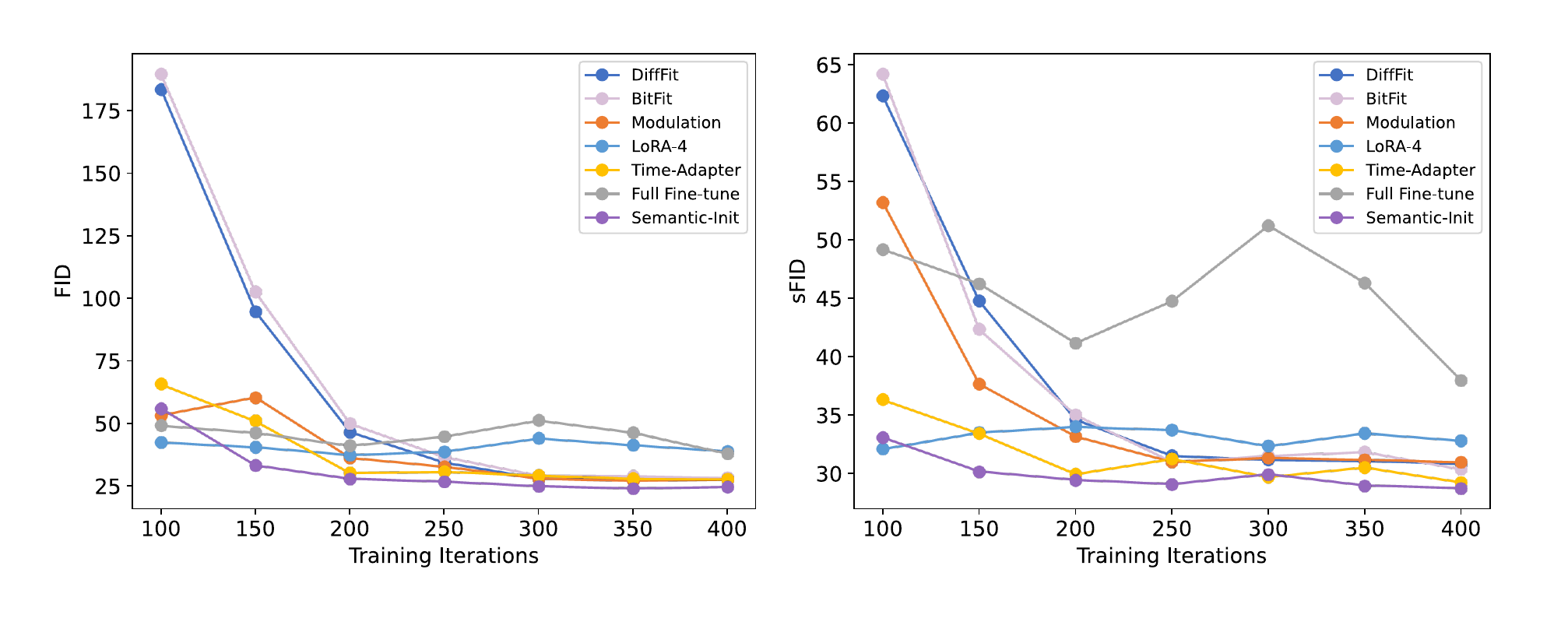}
  \caption{The generative performance of different fine-tuning methods in early training.}
  \label{fig:convergence}
\end{figure}

\begin{table}[tb]
  \setlength{\tabcolsep}{8pt}
  \caption{Ablation experiments on different modules of DiTs on Ithaca365 dataset. Our proposed DriveDiTFit selects MHSA and $MLP_c$  as fine-tuning modules.
  }
  \label{tab:ablations}
  \centering
  \begin{tabular}{@{}cccccc@{}}
    \toprule
    MHSA       & $MLP_b$           & $MLP_c$       & Patch Conv & FID$\downarrow$ & Params(M)                 \\ \midrule
    \checkmark &                   &               &            & 30.89           & 1.081(0.16\%)             \\
    \checkmark &                   & \checkmark    &            & 27.45           & 2.372(\underline{0.35\%}) \\
    \checkmark & \checkmark        & \checkmark    &
               & \underline{27.09} & 3.984(0.58\%)                                                            \\
    \checkmark & \checkmark        & \checkmark    & \checkmark & 29.84           & 3.991(0.59\%)             \\ \bottomrule
  \end{tabular}
\end{table}

\begin{table}[tb]
  \setlength{\tabcolsep}{8pt}
  \caption{Ablation experiments on DriveDiTFit's components. Each component helps improve the quality and fidelity of the generated samples.
  }
  \label{tab:ablations_modulation}
  \centering
  \begin{tabular}{@{}ccccccccc@{}}
    \toprule
    Modulation & SSEI       & OSL        & RoPE       & PT         & FID$\downarrow$   & sFID$\downarrow$  & Prec.$\uparrow$   & Recall $\uparrow$ \\ \midrule
    \checkmark &            &            &            &            & 27.45             & 31.21             & 0.414             & 0.764             \\
    \checkmark & \checkmark &            &            &            & 26.93             & 29.10             & 0.448             & 0.760             \\
    \checkmark & \checkmark & \checkmark &            &            & 25.35             & 29.29             & 0.488             & 0.731             \\
    \checkmark & \checkmark & \checkmark & \checkmark &            & 23.80             & 29.20             & 0.481             & 0.742             \\
    \checkmark & \checkmark & \checkmark & \checkmark & \checkmark & \underline{18.64} & \underline{26.13} & \underline{0.626} & \underline{0.795} \\ \bottomrule
  \end{tabular}
\end{table}

\textbf{Ablation studies.}  We first explore the effects of fine-tuning various modules of DiTs on their generative performance. As seen in \cref{tab:ablations_modulation}, fine-tuning the Multi-Head Self-Attention (MHSA) modules and the condition MLP ($MLP_c$) significantly enhances performance. This may stem from the substantial gap between the driving dataset and the original training dataset, which likely impact attention mechanisms and the utilization of conditional embeddings. Conversely, adjusting the MLP layer within the transformer blocks ($MLP_b$) does not markedly improve generation performance and only adds extra parameters. Hence, fine-tuning this module is not within the scope of our subsequent considerations. We also observed a decrease in model performance upon fine-tuning the Patch Convolution (Patch Conv) module. We hypothesize that the method of mapping $2\times2$ patches into a high-dimensional space remains effective for the driving dataset, as the method for mapping small patches, learned on the large-scale ImageNet dataset, demonstrates strong generalizability.

The findings depicted in \cref{fig:convergence} demonstrate that semantic-selective embedding initialization outperforms other fine-tuning methods by achieving better FID and sFID in  150 tuning iterations. This indicates that our approach enables DiT models to converge more rapidly and attain superior generative capabilities. Note that since the VPT method is not suited for zero initialization or identity initialization, resulting in slow prompt learning, we do not plot its training curve in the graph. We also examine the category embeddings from ImageNet corresponding to different conditions in the Ithaca365 dataset, within the semantic encoding space of CLIP. For example,  the category most closely aligned with snowy conditions is snowplow, characterized by backgrounds of expansive white roads, mirroring those found in snowy scenarios.  This similarity in the semantic space suggests that embeddings with close resemblance facilitate a quicker adaptation from the source to the target distribution, requiring less fine-tuning time and thus enhancing generative performance.

We conduct further ablation studies on the DriveDiTFit method, with results detailed in \cref{tab:ablations}. Each component contributes to improve the quality and fidelity of the generated samples. It is worth noting that the implementation of a Scos schedule significantly boosts generative performance by delaying the point at which vehicle details are obscured by noise. This approach provides an extended time frame for detailed generation during the denoising process. Importantly, this progressive fine-tuning strategy is not limited to DiT models but is also applicable to the fine-tuning of other diffusion models where the original noise schedule is suboptimal.

\subsection{Qualitative  Evaluation}
To gain a comprehensive understanding of the DriveDiTFit approach, we showcase its ability to generate synthesized images across a variety of weather and lighting conditions, alongside demonstrating the effectiveness of progressive tuning strategies employing a Scos schedule.

\begin{figure}[tb]
  \centering
  \setlength{\abovecaptionskip}{0.2cm}
  \includegraphics[width=1.0\textwidth]{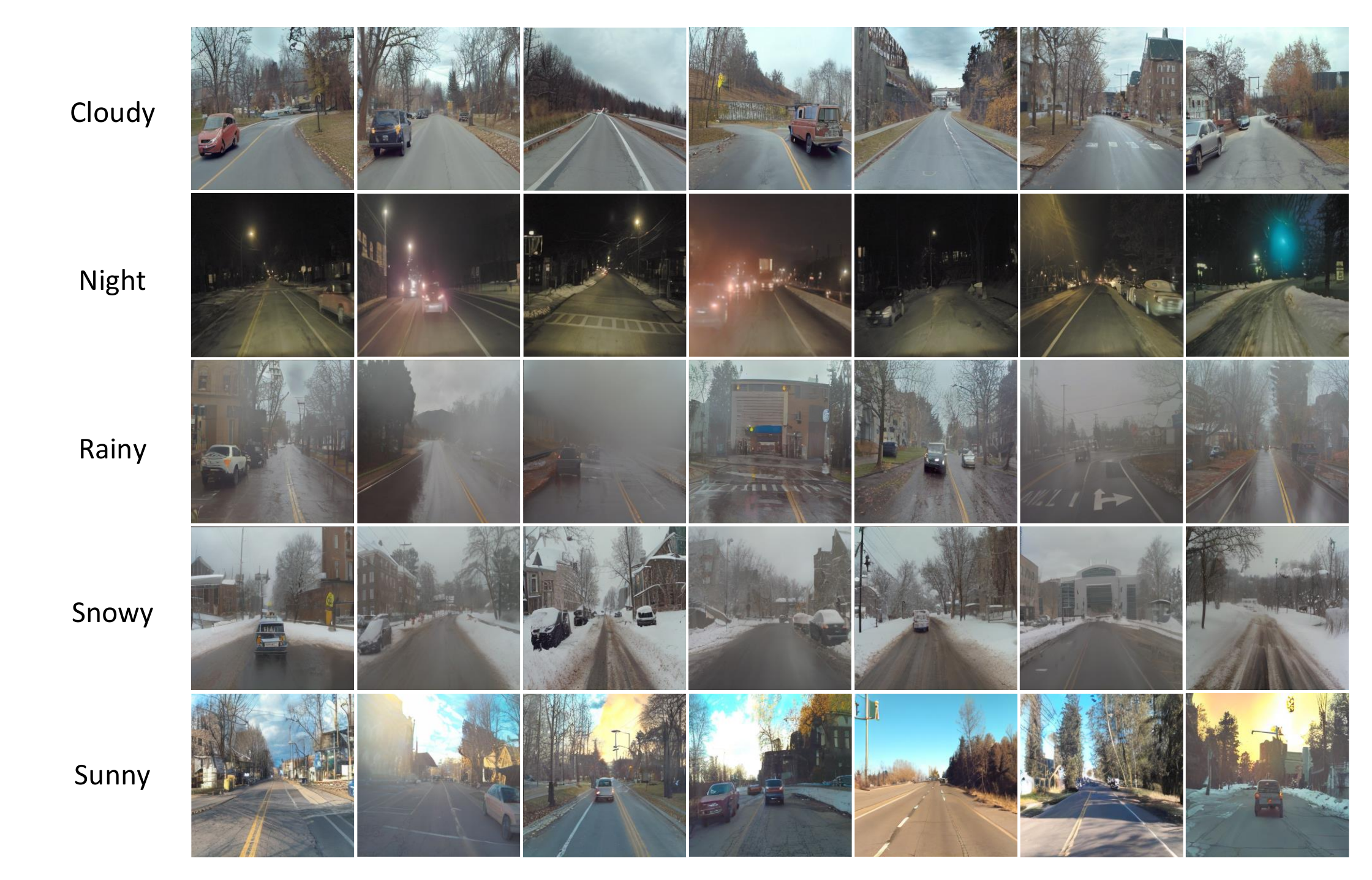}
  \caption{Samples from DiTs fine-tuned on Ithaca365 datasets through DriveDiTFit method, each with a resolution of 256 $\times$ 256.}
  \vspace{-15pt}
  \label{fig:generated}
\end{figure}

As illustrated in \cref{fig:generated}, the DiT model, once fine-tuned with DriveDiTFit, can generate high-quality and huge-diversity driving scenario images under given conditional embeddings. Remarkably, this model can create realistic depictions of vehicles on roads without relying on structured conditional inputs like vehicle bounding boxes\cite{drivemodel1} or segmentation maps\cite{drivemodel2}. Beyond learning evident weather or lighting characteristics, the fine-tuned model adeptly captures complex physical details, such as puddle reflections during rain, road shadows in sunny conditions, snow coverage, and the glow of vehicle lights at night. These observations demonstrate the reliability of our proposed fine-tuning method.

\begin{figure}[tb]
  \centering
  \setlength{\abovecaptionskip}{0.2cm}
  \includegraphics[width=1.0\textwidth]{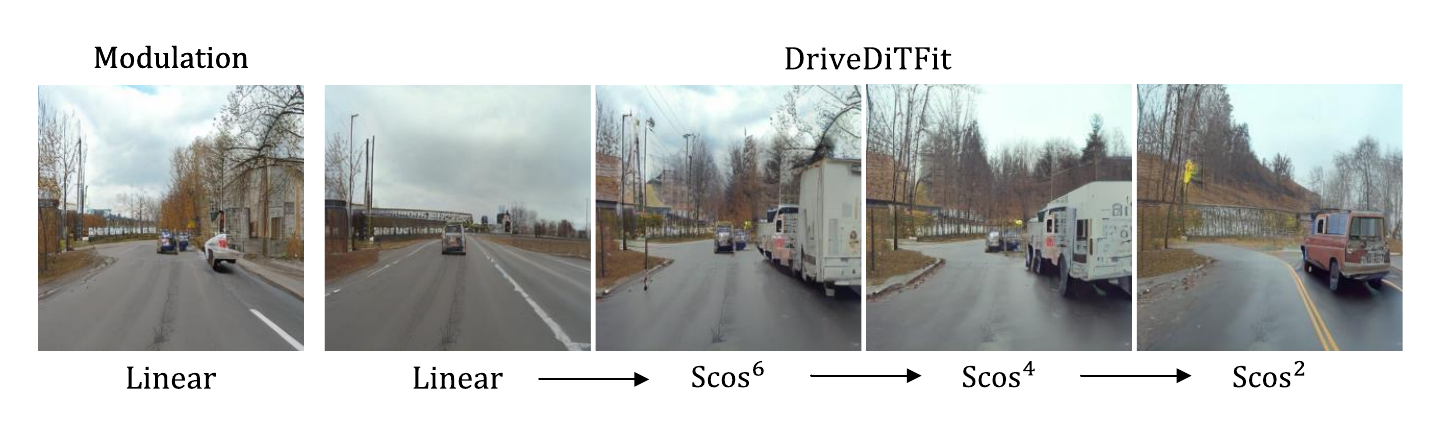}
  \caption{The process visualization of progressive tuning strategies with Scos schedule.}
  \vspace{-15pt}
  \label{fig:process}
\end{figure}

We further present the impact of the DriveDiTFit components, with a particular focus on the generation of details at various phases of the progressive tuning process. This visualization serves to validate the suitability of the Scos schedule for driving datasets. Employing a consistent noise, the integration of vehicle masks and the RoPE module, which leverages local attention mechanisms, significantly contributes to the accurate formation of complete vehicle contours, in contrast to traditional modulation methods. Furthermore, the use of the Scos schedule in the progressive tuning regimen notably refines the representation of intricate details, methodically enhancing the visualization of complex features such as tires, lane markings, and windows, thereby demonstrating the effectiveness of this approach in producing highly detailed and realistic images.

\subsection{Downstream Task Evaluation}

To validate that the generated dataset benefits downstream perception tasks in autonomous driving applications, particularly in enhancing the performance of visual perception models for imbalanced categories (such as adverse weather and lighting conditions), we choose the object detection task using the YOLOv8 model. We train the model separately on a real dataset and a mixed dataset (real + generated data) using default training parameters on a single 4090 GPU. The trained object detection models are then evaluated on a validation set consisting of adverse weather and lighting conditions, including rainy, nighttime, and snowy environments. As shown in \cref{tab:downtask}, compared to the real dataset, the generated dataset improved the YOLOv8 model's performance by 0.021 on the Ithaca365 dataset and by 0.019 on the BDD100K dataset. This demonstrates that the proposed method's generated dataset can enhance the YOLOv8 model's performance in object detection tasks, further confirming that the generated dataset is beneficial for downstream tasks.

\begin{table}[t]
  \setlength{\tabcolsep}{8pt}
  \caption{Validation results for the effectiveness of the generated dataset in the object detection task.
  }
  \label{tab:downtask}
  \centering
  \begin{tabular}{@{}ccccc@{}}
    \toprule
    \multirow{2}{*}{Datasets} & \multicolumn{2}{c}{Ithaca365} & \multicolumn{2}{c}{BDD100K}                             \\ \cmidrule(l){2-5}
                              & Real                          & Mixed                       & Real  & Mixed             \\ \midrule
    mAP@50                    & 0.502                         & \underline{0.523}           & 0.395 & \underline{0.414} \\ \bottomrule
  \end{tabular}
\end{table}

\section{Conclusion}
In this paper, we introduce DriveDiTFit, an innovative approach for efficient generation of autonomous driving data through fine-tuning pre-trained diffusion transformers. From the model architecture perspective, we have developed a modulation technique designed to adjust the overall style and layout of the generated samples, making it suitable for autonomous driving datasets that significantly diverge from the source dataset. Building on this, we have implemented a semantic-selective embedding initialization approach to help the model more rapidly learn the data distribution and achieve superior generative performance.  Notably, we are pioneers in identifying the influence of the original noise schedule on downstream datasets and have introduced a novel Scos schedule and an object-sensitive loss coupled with a progressive fine-tuning strategy, to enhance the detail generation of small objects in scenarios. Our experiments demonstrate that our proposed method can produce high-quality, diverse driving datasets under various weather and lighting conditions.

\label{sec:manuscript}

\bibliographystyle{splncs04}
\bibliography{egbib}

\end{document}